\def\year{2021}\relax
\documentclass[letterpaper]{article} 
\usepackage{aaai21}  
\usepackage{times}  
\usepackage{helvet} 
\usepackage{courier}  
\usepackage[hyphens]{url}  
\usepackage{graphicx} 
\usepackage[colorlinks,linkcolor=red]{hyperref}
\usepackage{natbib}  
\usepackage{caption} 
\usepackage{mathtools}
\usepackage{amsmath,amssymb}
\usepackage{multirow}
\usepackage{makecell}
\usepackage{xcolor}
\usepackage{pifont}
\newcommand{\etal}{\textit{et al}. }
\newcommand{\cmark}{\ding{51}}%
\title{Similarity Reasoning and Filtration for Image-Text Matching}
\author {
    Haiwen Diao,\textsuperscript{\rm 1}
    Ying Zhang,\textsuperscript{\rm 2}
    Lin Ma,\textsuperscript{\rm 3}
    Huchuan Lu\textsuperscript{\rm 1}\thanks{Corresponding author}\\
}
\affiliations {
    \textsuperscript{\rm 1} Dalian University of Technology, Dalian, China\\
    \textsuperscript{\rm 2} Tencent AI Lab, Shenzhen, China\\
    \textsuperscript{\rm 3} Meituan, Beijing, China\\
    r1228240468@mail.dlut.edu.cn, yinggzhang@tencent.com,\\
    lhchuan@dlut.edu.cn,
    forest.linma@gmail.com}

\begin{document}

\maketitle

\begin{abstract}
Image-text matching plays a critical role in bridging the vision and language, and great progress has been made by exploiting the global alignment between image and sentence, or local alignments between regions and words. However, how to make the most of these alignments to infer more accurate matching scores is still underexplored. In this paper, we propose a novel Similarity Graph Reasoning and Attention Filtration (SGRAF) network for image-text matching. Specifically, the vector-based similarity representations are firstly learned to characterize the local and global alignments in a more comprehensive manner, and then the Similarity Graph Reasoning (SGR) module relying on one graph convolutional neural network is introduced to infer relation-aware similarities with both the local and global alignments. The Similarity Attention Filtration (SAF) module is further developed to integrate these alignments effectively by selectively attending on the significant and representative alignments and meanwhile casting aside the interferences of non-meaningful alignments. We demonstrate the superiority of the proposed method with achieving state-of-the-art performances on the Flickr30K and MSCOCO datasets, and the good interpretability of SGR and SAF modules with extensive qualitative experiments and analyses.

\end{abstract}

\section{Introduction}
Image-text matching refers to measuring the visual-semantic similarity between image and text, which is becoming increasingly significant for various vision-and-language tasks, such as cross-modal retrieval \cite{SGM}, image captioning \cite{BU_TDA}, text-to-image synthesis \cite{attngan18}, and multimodal neural machine translation \cite{nmt17}. Although great progress has been made in recent years, image-text matching remains a challenging problem due to complex matching patterns and large semantic discrepancies between image and text.

\begin{figure}[htpb]
	\centering
	\begin{tabular}{@{}c}
		\includegraphics[width=1.0\linewidth, height=0.35\linewidth,trim=5 240 90 0,clip]{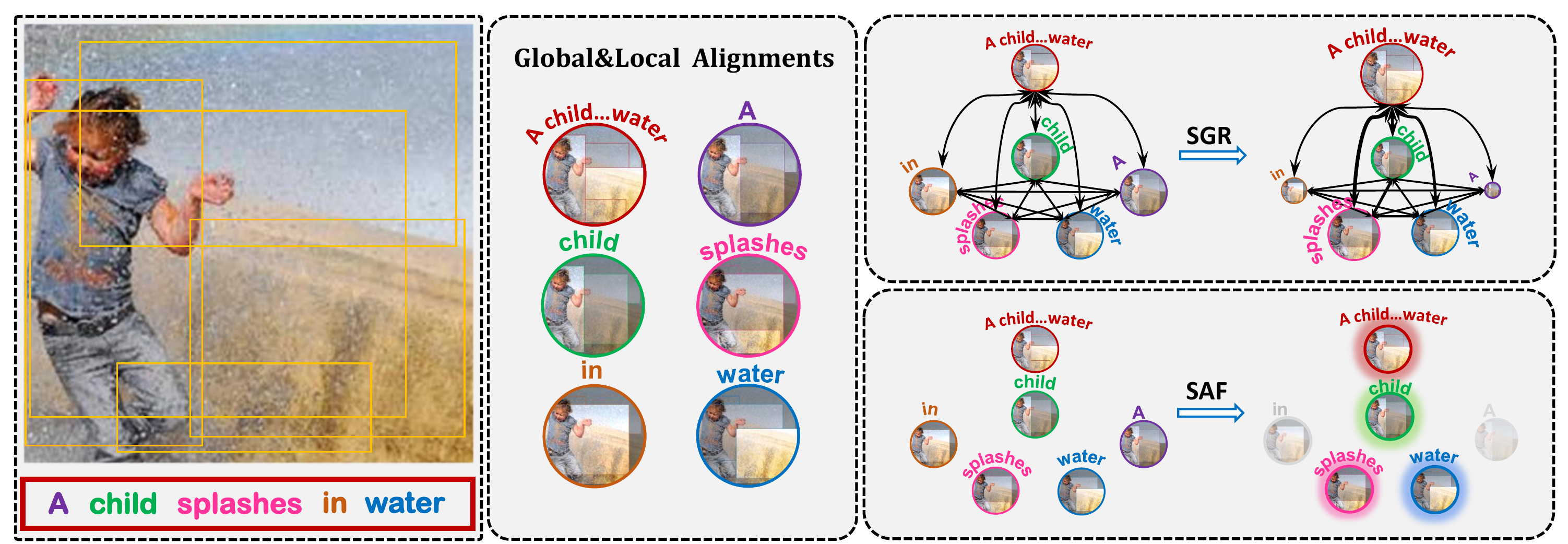} 
	\end{tabular}
	\caption{Illustration of the SGRAF. Nodes of red and other colors encode image-text and region-word alignments respectively. SGR module captures their relationships to achieve comprehensive similarity reasoning and SAF module reduces the interferences of less-meaningful alignments}
	\label{fig:motivation}
\end{figure}

To accurately establish the association between the visual and textual observations, a large proportion of methods \cite{RRF,DAN,SCAN,PVSE,CAMP,VSRN,SGM} utilize deep neural networks to firstly encode image and text into compact representations, and then learn to measure their similarity under the guidance of a matching criterion. For example, Wang \etal \cite{DSPE} and Faghri \etal \cite{VSE++} map the whole image and the full sentence into a common vector space, and compute the cosine similarity between the global representations. To improve the discriminative ability of the unified embeddings, many strategies such as semantic concept learning \cite{SCO,SCG} and region relationship reasoning \cite{VSRN} are developed to enhance visual features by incorporating local region semantics.
However, these approaches fail to capture the local interactions between image regions and sentence fragments, leading to limited interpretability and performance gains.
To address this problem, Karpathy \etal \cite{DVSA} and Lee \etal \cite{SCAN} propose to discover all the possible alignments between image regions and sentence fragments, which produce impressive retrieval results and inspire a surge of works \cite{CAMP,RDAN,CAAN,IMRAM,ACME} to explore more accurate fine-grained correspondence. Although noticeable improvements have been made by designing various mechanisms to encode more powerful features or capture more accurate alignments, these approaches neglect the importance of similarity computation, which is the key to explore the complex matching patterns between image and text.

To be more specific, there are three defects in previous approaches. Firstly, these methods compute scalar-based cosine similarities between local features, which may not be powerful enough to characterize the association patterns between regions and words. 
Secondly, most of them aggregate all the latent alignments between regions and words simply with max pooling \cite{DVSA} or average pooling \cite{SCAN,IMRAM}, which hinders the information communication between local and global alignments, and thirdly, fails to consider the distractions of less-meaningful alignments, such as the alignments built with \texttt{"a"} and \texttt{"in"}, as shown in Figure~\ref{fig:motivation}.

To address these problems, in this paper we propose a novel Similarity Graph Reasoning and Attention Filtration (SGRAF) network for image-text matching. Specifically, we start with capturing the global alignments between the whole image and the full sentence, as well as the local alignments between image regions and sentence fragments. Instead of characterizing these alignments with scalar-based cosine similarity, we propose to learn the vector-based similarity representations to model the cross-modal associations more effectively. Then we introduce the Similarity Graph Reasoning (SGR) module, which relies on a Graph Convolution Neural Network (GCNN) to reason more accurate image-text similarity via capturing the relationship between local and global alignments. Furthermore, we develop the Similarity Attention Filtration (SAF) module to aggregate all the alignments attended by different significance scores, which reduces the interferences of non-meaningful alignments and achieves more accurate cross-modal matching results. 
Our main contributions are summarized as follows: 
\begin{itemize}
	\item We propose to learn the vector-based similarity representations for image-text matching, which enables greater capacity in characterizing the global alignments between images and sentences, as well as the local alignments between regions and words.
	\item We propose the Similarity Graph Reasoning (SGR) module to infer the image-text similarity with graph reasoning, which can identify more complex matching patterns and achieve more accurate predictions via capturing the relationship between local and global alignments.
	\item We attempt to consider the interferences of non-meaningful words in similarity aggregation, and propose an effective Similarity Attention Filtration (SAF) module to suppress the irrelevant interactions for further improving the matching accuracy.
\end{itemize}

\section{Related Work}

\subsection{Image-Text Matching}

{\bf Feature Encoding} Many prior Approaches \cite{DVSA,PVSE,RRF,DAN,SCAN,CAMP,VSRN,SGM} focused on feature extraction and optimization for cross-modal retrieval. For textual features, Frome \etal \cite{DeViSE} employed Skip-Gram \cite{Skip-Gram} to extract word representations. Klein \etal \cite{GMM-FV} explored Fisher Vectors (FV) \cite{FV} for text representation. Kiros \etal \cite{GRU} adopted a GRU as the text encoder. For visual features, Liu \etal \cite{RRF} adapted Recurrent Residual networks to refine global embeddings. \cite{PVSE, MMCA} employed multi-head self-attention to combine global context with locally-guided features. Besides, Some works \cite{DAN, SAN} exploited block-based visual attention to gather semantics on feature maps, while \cite{SCAN,CAMP,PFAN,VSRN,SGM,DP-RNN} followed \cite{BU_TDA} to obtain region-based features of visual objects with the pre-trained model on Visual Genomes \cite{VisualGenome}. Especially, \cite{DP-RNN} explored Bi-GRU to gain high-level object features, while \cite{VSRN,SGM} proposed GCN-based networks to generate relationship-enhanced object features. We employ self-attention \cite{Self-Attention} on region or word features to get image or text representation. We concentrate on the similarity encoding mechanism that models global image-text and local region-word alignments comprehensively and fully encodes fine-grained relations between image and text.

{\bf Similarity Prediction} Most existing works \cite{VSE++,DSPE,Dualpath,Order,GXN} for image-text matching learned the joint embedding and the similarity measures for cross-modal matching. For global alignments, some works \cite{VSE++,DSPE,RRF,PVSE,DAN,VSRN} explored a joint space and calculated the inner product (e.g. cosine distance) for similarity computation. Others \cite{Order,GXN} introduced an ordered representations to measure antisymmetric visual-semantic hierarchy. For local alignments, most networks \cite{DVSA,SCAN,RDAN,PFAN,IMRAM} computed scalar-based alignments and adopted simple operation (e.g. sum and average) to fuse local alignments. For example, Lee \etal \cite{SCAN} studied the latent semantic alignments among region-words pairs and integrated local cosine alignments by average or LogSumExp. Differently, our network aggregates similarities by exploring global-local relationships among vector-based alignments and reducing the distraction from less-meaningful ones.

\subsection{Graph Convolution Network}
The researches based on Graph modeled the dependencies between concepts and facilitated graph reasoning such as GCNN \cite{CNG,GCN}, and Gated Graph Neural Network (GGNN) \cite{GNN}. These graph neural networks have been widely employed in various visual semantic tasks, such as image captioning \cite{AESG}, VQA \cite{GSR}, and grounding referring expressions \cite{LGGAN}. In recent years, there are several approaches to utilize graph structures to enhance single visual or textual features referring to image-text matching. Shi \etal \cite{SCG} adopted Scene Concept Graph (SCG) by using image scene graphs and frequently co-occurred concept pairs as scene common-sense knowledge. Li \etal \cite{VSRN} proposed Visual Semantic Reasoning to build up connections between image regions and generate visual representations with semantic relationships. Wang \etal \cite{SGM} employed visual scene graph and textual scene graph, each of which separately refines visual and textual features including objects and relationships. They all focus on "feature encoding" by learning single-modality contextualized representations, while our SGR targets at "similarity reasoning" and explores more complex matching patterns with global and local cross-modal alignments.

\subsection{Attention Mechanism}
The attention mechanism has been applied to adaptively filter and aggregate information in natural language processing. When it comes to image-text matching, it has been intended to attend to certain parts of visual and textual data. \cite{SCAN,PFAN} developed Stacked Cross Attention to match latent alignments using both image regions and textual words as context. \cite{BFAN,RDAN,CAMP} designed more complicated Cross Attentions to improve image-text matching. Chen \etal \cite{IMRAM} proposed an Iterative Matching with Recurrent Attention Memory to explore fine-grained region-word correspondence progressively. We adopt textual-to-visual attention \cite{SCAN} with region-word pairs and calculate textual-attended alignments. In this paper, our SAF aims to discard less-semantic alignments instead of exploiting precise cross-modal attention.

\section{Method}
\begin{figure*}[ht]
	\centering
	\begin{tabular}{@{}c}
		\includegraphics[width=1.0\linewidth, height=0.33\linewidth,trim= 10 270 130 0,clip]{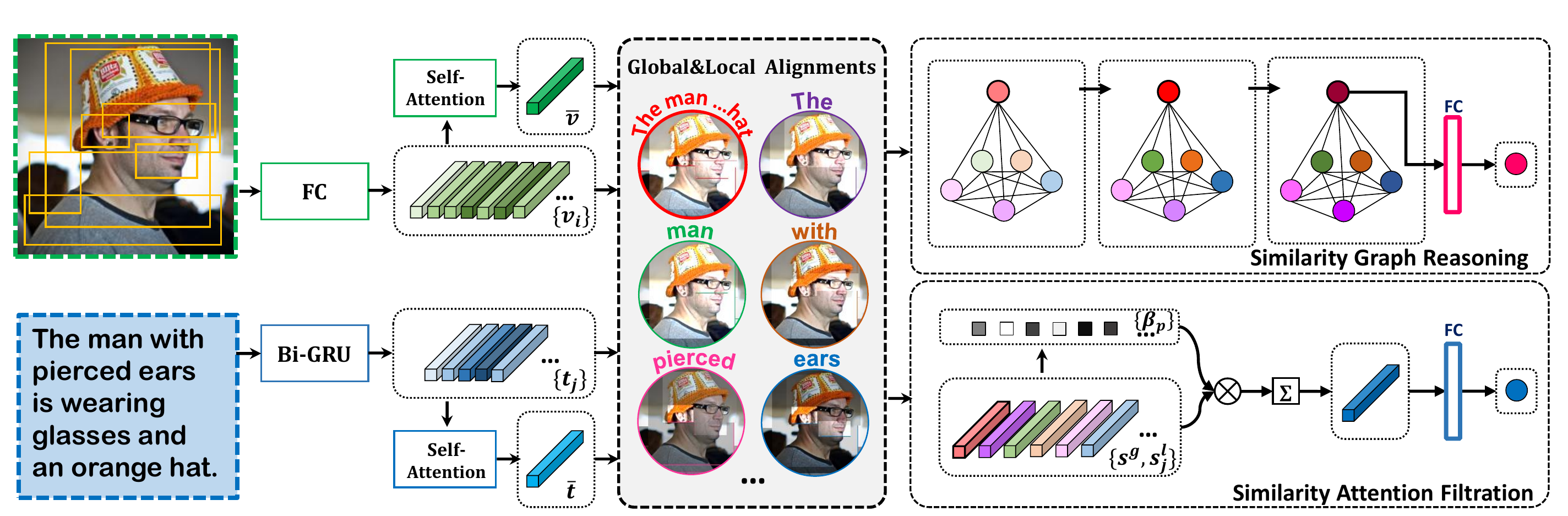} 
	\end{tabular}
	\caption{The proposed SGRAF network for image-text matching. The image and sentence are firstly encoded into local and global feature representations, followed by a similarity representation computation module to capture the correspondence between all local and global cross-modal pairs. The Similarity Graph Reasoning (SGR) module reasons the similarity with giving consideration to the relationship between all the alignments, and the Similarity Attention Filtration (SAF) module attends to more informative alignments for more accurate similarity prediction} \label{fig:framework}
\end{figure*}

In this section, we focus on improving the visual-semantic similarity learning via capturing the relationship between local and global alignments, and suppressing the disturbance of less-meaningful alignments. As illustrated in Figure \ref{fig:framework}, we begin with introducing how to encode the visual and textual observations, and then compute the similarity representations of all local and global representation pairs. Afterwards, we elaborate on the proposed Similarity Graph Reasoning (SGR) module for relation-aware similarity reasoning and Similarity Attention Filtration (SAF) module for representative similarity aggregation. Finally, we present the detailed implementations of training objectives and inference strategies with both the SGR and SAF modules. 

\subsection{Generic Representation Extraction}
\label{secIRM}
\subsubsection{Visual Representations.} 
For each input image, we follow \cite{BU_TDA} to extract $K$ region-level visual features, with the Faster R-CNN \cite{FasterR-CNN} model pre-trained on Visual Genomes \cite{VisualGenome}. We add a fully-connect layer to transform them into $d$-dimensional vectors as local region representations $\boldsymbol{V} = \{\boldsymbol{v}_{1},...,\boldsymbol{v}_{K}\}$, with $\boldsymbol{v}_{i}\in \mathbb{R}^{d}$. Afterwards, we perform self-attention mechanism \cite{Self-Attention} over the local regions, which adopts average feature $\bar{\boldsymbol{q}_{v}} = \frac{1}{K} \sum_{i=1}^{K}\boldsymbol{v}_{i}$ as the query and aggregates all the regions to obtain global representation $\bar{\boldsymbol{v}}$.

\subsubsection{Textual Representations.}
Given a sentence, we split it into $L$ words with tokenization technique, and sequentially feed the word embeddings into a bi-directional GRU \cite{Bi-GRU}. The representation of each word is then obtained by averaging the forward and backward hidden state at each time step,  with $\boldsymbol{T} = \{\boldsymbol{t}_{1},...,\boldsymbol{t}_{L}\}$, and $\boldsymbol{t}_{j}\in \mathbb{R}^{d}$ denoting the representation of $j$-th word. Similarly, the global text representation $\bar{\boldsymbol{t}}$ is computed by the self-attention method over all the word features. 

\subsection{Similarity Representation Learning}
\subsubsection{Vector Similarity Function.} Most previous methods utilize the cosine or Euclidean distance to represent the similarity between two feature vectors, which can capture the relevance to a certain degree while lacks the detailed correspondence. In this paper, we compute a similarity representation, which is a similarity vector instead of a similarity scalar, to capture more detailed associations between feature representations from different modalities. The similarity function between vector $\boldsymbol{x} \in \mathbb{R}^{d}$ and  $\boldsymbol{y}  \in \mathbb{R}^{d}$ is defined as
\begin{equation}
\label{eq:sfunc}
\boldsymbol{s}(\boldsymbol{x}, \boldsymbol{y}; \boldsymbol{W}) = \frac{\boldsymbol{W}{\left | \boldsymbol{x}- \boldsymbol{y} \right |}^{2}}{{\left \| \boldsymbol{W}{\left | \boldsymbol{x}- \boldsymbol{y} \right |}^{2} \right \|}_{2}}
\end{equation}
where ${\left | \cdot  \right |}^{2}$ and ${\left \| \cdot \right \|}_{2}$ indicate element-wise square and $\ell_2$-norm respectively, and $\boldsymbol{W}\in \mathbb{R}^{m\times d}$ is a learnable parameter matrix to obtain the $m$-dimensional similarity vector.

\label{secSRM}
\subsubsection{Global Similarity Representation.} We compute the similarity representation between 
the global image feature $\bar{\boldsymbol{v}}$ and sentence features $\bar{\boldsymbol{t}}$ with Eq.~\eqref{eq:sfunc},
\begin{equation}
\label{eq:sglo}
\boldsymbol{s}^{g} = \boldsymbol{s}(\bar{\boldsymbol{v}}, \bar{\boldsymbol{t}}; \boldsymbol{W_g})
\end{equation}
where $\boldsymbol{W}_{g}\in \mathbb{R}^{m\times d}$ aims to learn the global similarity  representation.

\subsubsection{Local Similarity Representation.}  To exploit local similarity representations between local features of visual and textual observations, we apply textual-to-visual attention \cite{SCAN} to attend on each region with respect to each word. Attention weight for each region is computed by
\begin{equation}
{\alpha}_{ij}=\frac{exp(\lambda\hat{c}_{ij})}{{\sum}_{i=1}^{K}exp(\lambda\hat{c}_{ij})}
\end{equation}
Here the weight ${\alpha}_{ij}$ is calculated by the softmax function with a temperature parameter $\lambda$. $c_{ij}$ indicates the cosine similarity between region feature $\boldsymbol{v}_{i}$ and word feature $\boldsymbol{t}_{j}$, ${\hat{c}_{ij}} = {\left [ c_{ij} \right ]}_{+}/\sqrt{{\sum}_{j=1}^{L}{\left [ c_{ij} \right ]}_{+}^{2}}$ aims to normalize the cosine similarity matrix, and $\left [ x \right ]_{+} = max(x, 0)$.

Then we generate the attended visual features $\boldsymbol{a}_{j}^{v}$  with respect to $j$-th word by
\begin{equation}
\boldsymbol{a}_{j}^{v} = \sum \limits_{i=1}^{K}{\alpha}_{ij}\boldsymbol{v}_{i},
\end{equation}
and finally we compute the local similarity representation between $\boldsymbol{a}_{j}^{v}$  and $\boldsymbol{t}_{j}$ as 
\begin{equation}
\label{eq:sloc}
\boldsymbol{s}_{j}^{l}= \boldsymbol{s}(\boldsymbol{a}_{j}^{v}, \boldsymbol{t}_{j}; \boldsymbol{W}_{l} )
\end{equation}
where $\boldsymbol{W}_{l}\in \mathbb{R}^{m\times d}$ is also a learnable parameter matrix.
The local similarity representations capture the associations between a specific word and its corresponding image regions, which exploit more fine-grained visual-semantic alignments to boost the similarity prediction.

\subsection{Similarity Graph Reasoning}
\label{secSGR}
\subsubsection{Graph Building.} To achieve more comprehensive similarity reasoning, we build a similarity graph to propagate similarity messages among the possible alignments at both local and global levels. More specifically, we take all the word-attended similarity representations and the global similarity representation as graph nodes, i.e.  $\mathcal{N} =\{\boldsymbol{s}_{1}^l, ....,  \boldsymbol{s}_{L}^l, \boldsymbol{s}^g\}$,  and follow \cite{GRN19} to compute the edge from node $\boldsymbol{s}_{q} \in \mathcal{N}$  to $\boldsymbol{s}_{p} \in \mathcal{N}$ as
\begin{equation}
e(\boldsymbol{s}_{p}, \boldsymbol{s}_{q}; \boldsymbol{W}_{in}, \boldsymbol{W}_{out})=\frac{exp((\boldsymbol{W}_{in}{\boldsymbol{s}}_{p})(\boldsymbol{W}_{out}{\boldsymbol{s}}_{q}))}{{\sum}_{q}exp((\boldsymbol{W}_{in}{\boldsymbol{s}}_{p})(\boldsymbol{W}_{out}{\boldsymbol{s}}_{q}))},
\end{equation}
where $\boldsymbol{W}_{in}\in \mathbb{R}^{m\times m}$ and $\boldsymbol{W}_{out}\in \mathbb{R}^{m\times m}$ are the linear transformations for incoming and outgoing nodes, respectively. Note that the edges between node $\boldsymbol{s}_{p}$ and $\boldsymbol{s}_{q}$ are directed, which allow efficient and complex information propagation for similarity reasoning.

\subsubsection{Graph Reasoning.} With the constructed graph nodes and edges, we perform similarity graph reasoning by updating the nodes and edges with 
\begin{equation}
\hat{\boldsymbol{s}}_{p}^{n} = \sum \nolimits_{q} e(\boldsymbol{s}_{p}^{n}, \boldsymbol{s}_{q}^{n}; \boldsymbol{W}_{in}^n, \boldsymbol{W}_{out}^n)  \cdot \boldsymbol{s}_{q}^{n}
\end{equation}
\begin{equation}
\boldsymbol{s}_{p}^{n+1} = ReLU(\boldsymbol{W}_{r}^n \hat{\boldsymbol{s}}_{p}^{n}) 
\end{equation}
with $\boldsymbol{s}_{p}^{0}$ and  $\boldsymbol{s}_{q}^{0}$ taken from $\mathcal{N}$ at step $n=0$, and $\boldsymbol{W}_{r}^n$, $\boldsymbol{W}_{in}^n$, $\boldsymbol{W}_{out}^n$ are learnable parameters in each step. After current step of graph reasoning, the node $\boldsymbol{s}_{p}^{n}$ is replaced with $\boldsymbol{s}_{p}^{n+1}$.

We iteratively reason the similarity for $N$ steps, and take the output of the global node at the last step as the reasoned similarity representation, and then feed it into a fully-connect layer to infer the final similarity score. The SGR module enables the information propagation between local and global alignments, which can capture more comprehensive interactions to facilitate the similarity prediction.

\subsection{Similarity Attention Filtration}
\label{secSAF}
Although the exploitation of local alignments can boost the matching performance via discovering more fine-grained correspondence between image regions and sentence fragments, we notice that the less-meaningful alignments hinder the distinguishing ability when aggregating all the possible alignments in an undifferentiated way. Therefore we propose a Similarity Attention Filtration (SAF) module to enhance important alignments, as well as suppress ineffectual alignments, such as the alignments with \texttt{"the"}, \texttt{"be"} and etc.

Given the local and global similarity representations, we calculate an aggregation weight ${\beta}_{p}$ for each similarity representation $\boldsymbol{s}_{p} \in \mathcal{N}$ by
\begin{equation}
{\beta}_{p} = \frac{\delta(BN(\boldsymbol{W}_{f}{\boldsymbol{s}}_{p}))}{{\sum}_{\boldsymbol{s}_{q} \in \mathcal{N}}\delta(BN(\boldsymbol{W}_{f}{\boldsymbol{s}}_{q}))}
\end{equation}
where $\delta(\cdot)$ is the Sigmoid function, $BN$ indicates the batch normalization, and $\boldsymbol{W}_{f}\in \mathbb{R}^{m\times 1}$ is a linear transformation.

Then we aggregate the similarity representations with $\boldsymbol{s}_{f}=\sum_{\boldsymbol{s}_{p} \in \mathcal{N}} {\beta}_{p}  \boldsymbol{s}_{p} $, and feed $\boldsymbol{s}_{f}$ into a fully-connect layer to predict the final similarity between the input image and sentence. The SAF module learns the significance scores to increase the contribution of  more-informative similarity representations and meanwhile reduce the disturbance of less-meaningful alignments.

\setlength{\tabcolsep}{4pt}
\begin{table*}[ht]
	\begin{center}
		\renewcommand{\arraystretch}{1.1}
		\begin{tabular}{l|cccccc|cccccc}
			\hline
			\multirow{3}{*}{\bf Methods} &\multicolumn{6}{c} {MSCOCO dataset} &\multicolumn{6}{|c} {Flickr30K dataset} \\
			\;  &\multicolumn{3}{c}  {Sentence Retrieval} & \multicolumn{3}{c}  {Image Retrieval} &\multicolumn{3}{|c}  {Sentence Retrieval} & \multicolumn{3}{c}  {Image Retrieval}	\\
			\;  & R@1 & R@5 & R@10 & R@1 & R@5 & R@10 & R@1 & R@5 & R@10 & R@1 & R@5 & R@10\\
			\hline
			CAMP \cite{CAMP} &72.3 &94.8 &98.3 &58.5 &87.9 &95.0&68.1 &89.7 &95.2 &51.5 &77.1 &85.3\\
			SCAN \cite{SCAN} &72.7 &94.8 &98.4 &58.8 &88.4 &94.8&67.4 & 90.3 &95.8 &48.6 & 77.7 &85.2\\
			SGM \cite{SGM} &73.4 &93.8 &97.8 &57.5 &87.3 &94.3 &71.8 &91.7 &95.5 &53.5 &79.6 &86.5\\
			VSRN* \cite{VSRN} &74.0 &94.3 &97.8 &60.8 &88.4 &94.1 &70.4 &89.2 &93.7 &53.0 &77.9 &85.7\\
			RDAN \cite{RDAN} &74.6 &96.2 &98.7 &61.6 &89.2 &94.7&68.1 &91.0 &95.9 &54.1 &80.9 &87.2\\
			MMCA \cite{MMCA} &74.8 &95.6 &97.7 &61.6 &89.8 &95.2&74.2 &92.8 &96.4 &54.8 &81.4 &87.8\\
			BFAN \cite {BFAN} &74.9 &95.2 &- &59.4 &88.4 &-&68.1 &91.4 &- &50.8 &78.4 &-\\
			CAAN \cite{CAAN} &75.5 &95.4 &98.5 &61.3 &89.7 &95.2 &70.1 &91.6 &97.2 &52.8 &79.0 &87.9\\
			DPRNN \cite{DP-RNN} &75.3 &95.8 &98.6 &62.5 &89.7 &95.1 &70.2 &91.6 &95.8 &55.5 &81.3 &88.2\\
			PFAN \cite{PFAN} &76.5 &{\bf96.3} &{\bf99.0} &61.6 &89.6 &95.2 &70.0 &91.8 &95.0 &50.4 &78.7 &86.1\\
			VSRN \cite{VSRN} &76.2 &94.8 &98.2 &62.8 &89.7 &95.1&71.3 &90.6 &96.0 &54.7 &81.8 &88.2\\
			IMRAM \cite{IMRAM} &76.7 &95.6 &98.5 &61.7 &89.1 &95.0&74.1 &93.0 &96.6 &53.9 &79.4 &87.2\\
			\hline
			{\bf Ours(SAF)} &76.1 &95.4 &98.3 &61.8 &89.4 &95.3 &73.7 &93.3 &96.3 &56.1 &81.5 &88.0\\
			{\bf Ours(SGR)} &78.0 &95.8 &98.2 &61.4 &89.3 &95.4 &75.2 &93.3 &96.6 &56.2 &81.0 &86.5\\
			{\bf Ours(SGRAF}) &{\bf79.6} &96.2 &98.5 &{\bf63.2} &{\bf90.7} &{\bf96.1} &{\bf77.8} &{\bf94.1} &{\bf97.4} &{\bf58.5} &{\bf83.0} &{\bf88.8}\\
			\hline
		\end{tabular}
		\caption{Comparison of bi-directional retrieval results (R@K(\%)) on MSCOCO 1K test set and Flickr30K test set. VSRN* denotes a single model for a fair comparison with SGR. SGRAF denotes the whole framework with independent training}\label{tabcocof30k}
	\end{center}
\end{table*}

\subsection{Training Objectives and Inference Strategies}
\label{secSLF}
We utilize the bidirectional ranking loss \cite{VSE++} to train both the SGR and SAF modules. Given a matched image-text pair $(\boldsymbol{v}, \boldsymbol{t})$, and the corresponding hardest negative image  $\boldsymbol{v}^-$ and the hardest negative text $\boldsymbol{t}^-$ within a minibatch, we compute the bidirectional ranking loss  with 
\begin{equation}
\begin{split}
\mathcal{L}_r(\boldsymbol{v}, \boldsymbol{t}) = [ \gamma - \mathcal{S}_r(\boldsymbol{v}, \boldsymbol{t}) +  \mathcal{S}_r(\boldsymbol{v}, \boldsymbol{t}^-)]_{+} \\
+ [ \gamma - \mathcal{S}_r(\boldsymbol{v}, \boldsymbol{t}) +  \mathcal{S}_r(\boldsymbol{v}^-, \boldsymbol{t})]_{+} 
\end{split}
\end{equation}
where $\gamma$ is the margin parameter and $\mathcal{S}_r(\cdot, \cdot)$ indicates similarity prediction function implemented with SGR. Similarly, we define the training objectives on SAF module as $\mathcal{L}_f$.

In this paper, we explore different training and inference strategies with the proposed SGR and SAF modules:  joint training and independent training. For joint training, we combine $\mathcal{L}_r$ and $\mathcal{L}_f$ to train SGR and SAF modules simultaneously, where the similarity representations are shared for the proposed two modules. For independent training, we train the SGR and SAF modules separately. At the inference stage, we average the similarities predicted by SGR and SAF modules for the retrieval evaluation. 

\section{Experiments}
To verify the effectiveness of the our model, in this section we demonstrate extensive experiments on two benchmark datasets. We also introduce detailed implementations and training strategy of the proposed SGRAF model.

\subsection{Datasets and Settings}

\subsubsection{Datasets.}
We evaluate our model on the MSCOCO~\cite{MSCOCO} and Flickr30K~\cite{Flickr30k} datasets. The MSCOCO dataset contains 123,287 images, and each image is annotated with 5 annotated captions. The dataset is split into 113,287 images for training, 5000 images for validation and 5000 images for testing. We report results by averaging over 5 folds of 1K test images and testing on the full 5K images. The Flickr30K dataset contains 31,783 images with 5 corresponding captions each. Following the split in \cite{DeViSE}, we use 1,000 images for validation, 1,000 images for testing and the rest for training. 

\setlength{\tabcolsep}{2pt}
\begin{table}[ht]
	\begin{center}
		\renewcommand{\arraystretch}{1.1}
		\begin{tabular}{lcccc}
			\hline
			\multirow{2}{*}{\bf Methods} & \multicolumn{2}{c}  {Sen. Ret.} & \multicolumn{2}{c}  {Ima. Ret.} 	\\
			\;  & R@1  & R@10  & R@1  & R@10  \\
			\hline
			SGM \cite{SGM} &50.0 &87.9 &35.3 &76.5\\
			CAMP \cite{CAMP} &50.1 &89.7 &39.0 &80.2\\
			VSRN* \cite{VSRN} &50.3 &87.9 &37.9 &79.4\\
			SCAN \cite{SCAN} &50.4 &90.0 &38.6 &80.4\\
			CAAN \cite{CAAN} &52.5 &90.9 &41.2 &{\bf82.9}\\
			VSRN \cite{VSRN} &53.0 &89.4 &40.5 &81.1\\
			IMRAM \cite{IMRAM} &53.7 &91.0 &39.7 &79.8\\
			MMCA \cite{MMCA} &54.0 &90.7 &38.7 &80.8\\
			\hline
			{\bf Ours(SAF)} &53.3 &90.1 &39.8 &80.2\\
			{\bf Ours(SGR)} &56.9 &90.5 &40.2 &79.8\\
			{\bf Ours(SGRAF}) &{\bf57.8} &{\bf91.6} &{\bf41.9}  &81.3\\
			\hline
		\end{tabular}
		\caption{Comparison of bi-directional retrieval results (R@K(\%)) on MSCOCO 5K test set}
		\label{tabcoco5k}
	\end{center}
\end{table}

\subsubsection{Protocols.}
For image-text retrieval, we measure the performance by Recall at K (R@K) defined as the proportion of queries whose ground-truth is ranked within the top $K$. We adopt R@1, R@5 and R@10 as our evaluation metrics.

\setlength{\tabcolsep}{2pt}
\begin{table}[ht]
	\begin{center}
		\begin{tabular}{ccccccccccc}
			\hline
			\multirow{2}{*}{model}&\multirow{2}{*}{GLO}&\multirow{2}{*}{LOC}&\multicolumn{4}{c}{Step}&\multicolumn{2}{c}{Sen. Ret.}&\multicolumn{2}{c}{Ima. Ret.}\\
			&&& 1&2&3&4&R@1&R@10&R@1&R@10\\
			\hline 
			1 & \cmark &&&&&&62.4&92.6&46.0&83.1\\
			2 && \cmark & \cmark&&&&71.8 &95.6 &52.1 &82.3\\
			3 && \cmark &&& \cmark &&73.6 &96.1 &54.3 &85.1\\
			4 & \cmark & \cmark & \cmark &&& &74.2 &96.3 &55.5 &86.0\\
			5 & \cmark & \cmark && \cmark && &75.3 &{\bf96.7} &56.0 &85.9\\
			6 & \cmark & \cmark &&& \cmark &&75.2 &96.6 &{\bf56.2} &{\bf86.5}\\
			7 & \cmark & \cmark &&&& \cmark &{\bf76.2} &96.3 &55.0 &86.1\\
			\hline
		\end{tabular}
	\caption{The impact of SGR configurations. GLO and LOC respectively indicates the employment of global and local alignments, and Step denotes the graph reasoning steps}\label{tabSGR}
	\end{center}
\end{table}

\subsubsection{Implementation Details.}
For each image, we take the Faster-RCNN \cite{FasterR-CNN} detector with ResNet-101 provided by \cite{BU_TDA} to extract the top $K=36$ region proposals and obtain a 2048-dimensional feature for each region. For each sentence, we set the word embedding size as 300, and the number of hidden states as 1024. The dimension of similarity representation $m$ is 256, with smooth temperature $\lambda=9$, reasoning steps $N=3$, and margin $\gamma=0.2$.
Our model employs the Adam optimizer \cite{Adam} to train the SGRAF network with the mini-batch size of 128. The learning rate is set to be 0.0002 for the first 10 epochs and 0.00002 for the next 10 epochs on MSCOCO. For Flickr30K, we start training the SGR (SAF) module with learning rate 0.0002 for 30 (20) epochs and decay it by 0.1 for the next 10 epochs. We select the snapshot with the best performance on the validation set for testing.

\subsection{Quatitative Results and Analysis}

In this section, we present the retrieval results on the MSCOCO and Flickr30K datasets, aiming to demonstrate the effectiveness and superiority of the proposed approach.

\subsubsection{Comparisons on MSCOCO.}
Table \ref{tabcocof30k} and \ref{tabcoco5k} report the experimental results on MSCOCO dataset with 1K and 5K test images, separately. We can see that our proposed SGRAF model outperforms the existing methods, with the best R@1=$79.6\%$ for sentence retrieval and R@1=$63.2\%$ for image retrieval with 1K test images. 
For 5K test images, the proposed approach maintains the superiority with an improvement of more than $3\%$ on the R@1 results. It should be noted that competitive retrieval performance can be also achieved with the SGR/SAF module alone, demonstrating the effectiveness and complementarity of our modules. 

\subsubsection{Comparisons on Flickr30K.} 
Table \ref{tabcocof30k} compares the bidirectional retrieval results on Flickr30K dataset with the latest algorithms. We can observe that the SAF module alone produces comparable retrieval results and the SGR module achieves state-of-the-art performance with R@1 of $75.2\%$ and $56.2\%$ for sentence and image retrieval, separately. This verifies the effectiveness of exploiting the relationship between alignments to boost similarity reasoning. When we combine the SAF and SGR module, the performance is further improved to achieve the best R@1 of $77.8\%$ and $58.5\%$.

\subsection{Ablation Studies}

In this section, we carry a series of ablation studies to explore the impact of different configurations for the SGR module, the similarity representation learning module and the process of training. We also compare different strategies of similarity prediction to demonstrate the superiority of SGR and SAF modules. All the comparative experiments are conducted on the Flickr30K dataset.

\setlength{\tabcolsep}{1.5pt}
\begin{table}[ht]
	\begin{center}
		\begin{tabular}{ccccccccccccccc}
			\hline
			\multirow{2}{*}{\small{model}} &\multirow{2}{*}{{\small I2T}}&\multirow{2}{*}{{\small T2I}}&\multirow{2}{*}{{\small SS}}&\multirow{2}{*}{{\small SV}}&\multirow{2}{*}{{\small AA}}&\multirow{2}{*}{{\small SGR}}&\multirow{2}{*}{{\small SAF}} & \multicolumn{2}{c}{Sen. Ret.}&\multicolumn{2}{c}{Ima. Ret.} \\
			&&&&&&&&{\small R@1}&{\small  R@10}&{\small  R@1}&{\small R@10} \\
			\hline 
			1 & \cmark && \cmark && \cmark &&&66.7&94.1&43.2&82.3\\
			2 & \cmark &&& \cmark & \cmark &&&67.2&94.8&47.6&83.1\\
			3 & \cmark &&& \cmark && \cmark &&66.1&94.1&45.6&81.6\\
			4 & \cmark &&& \cmark &&& \cmark &{\bf68.2}&{\bf95.1}&{\bf49.8}&{\bf85.1}\\
			\hline
			5 && \cmark & \cmark && \cmark &&&62.6&93.6&45.3&82.4\\
			6 && \cmark && \cmark & \cmark &&&65.2&95.1&49.5&83.5\\
			7 && \cmark && \cmark && \cmark &&{\bf73.6} &96.1 &54.3 &85.1\\
			8 && \cmark && \cmark &&& \cmark &72.9 &{\bf96.3} &{\bf55.7} &{\bf87.8}\\
			\hline
		\end{tabular}
		\caption{The impact of Similarity configurations. I2T and T2I denotes the visual-to-textual and textual-to-visual attention to generate local similarity representations separately. SS denotes the scalar-based cosine similarity and SV indicates the vector-based similarity, and AA represents the average aggregation of all alignments}\label{tabFV}
	\end{center}
\end{table}
\setlength{\tabcolsep}{1.4pt}

\setlength{\tabcolsep}{1pt}
\begin{table}
	\begin{center}
		\begin{tabular}{ccccccccc}
			\hline
			\multirow{2}{*}{Dataset}&\multirow{2}{*}{SAF}&\multirow{2}{*}{SGR}&\multirow{2}{*}{Joint}&\multirow{2}{*}{Split}&\multicolumn{2}{c}{Sen. Ret.}&\multicolumn{2}{c}{Ima. Ret.}\\
			&&&&&R@1&R@10&R@1&R@10\\
			\hline 
			\multirow{4}{*}{MSCOCO}
			&\cmark&&&&76.1 &98.3 &61.8 &95.3\\
			&&\cmark&&&78.0 &98.2 &61.4 &95.4\\
			&\cmark&\cmark&\cmark& &77.8 &98.2 &61.6 &95.2\\
			&\cmark&\cmark&&\cmark&{\bf79.6} &{\bf98.5} &{\bf63.2} &{\bf96.1}\\
			\hline
			\multirow{4}{*}{Flickr30K}
			&\cmark&&&&73.7 &96.3 &56.1 &88.0\\
			&&\cmark&&&75.2 &96.6 &56.2 &86.5\\
			&\cmark&\cmark&\cmark& &75.1 &96.1 &56.2 &85.8\\
			&\cmark&\cmark&&\cmark&{\bf77.8} &{\bf97.4} &{\bf58.5} &{\bf88.8}\\
			\hline
		\end{tabular}
		\caption{The impact  of Training configurations on MSCOCO 1K test set and Flickr30K test set. Split and Joint denotes independent and joint training of two modules}\label{tabtrain}
	\end{center}
\end{table}

\subsubsection{Configurations of SGR module.}
In Table \ref{tabSGR} we investigate the effectiveness of each component in the SGR module. 1) Graph reasoning. We employ a framework without graph reasoning as the baseline(\#1), which adopts a fully-connected layer and sigmoid function on the global alignment to obtain the final similarity. Comparing \#1 and \#6 based on R@1, the SGR module achieves $12.8\%$ improvement for sentence retrieval and $10.2\%$ for image retrieval. 2) Reasoning steps setting. Comparing \#4, \#5, \#6 and \#7, we set the step of the SGR module to $3$ for maximum performance. 3) Global and local alignments. \#2 and \#3 only utilize local alignments for graph reasoning and adopt a mean-pooling operation on them after reasoning. Comparing \#2, \#4 and \#3, \#6, we discover that global similarity is beneficial for aggregating local similarities and exploring their relations which improves at least $1.6\%$ for sentence retrieval and $1.9\%$ for image retrieval on R@1.

\begin{figure*}[ht]
	\centering
	\begin{tabular}{@{}c}
		\includegraphics[width=1.0\linewidth, height=0.36\linewidth,trim=0 270 170 0,clip]{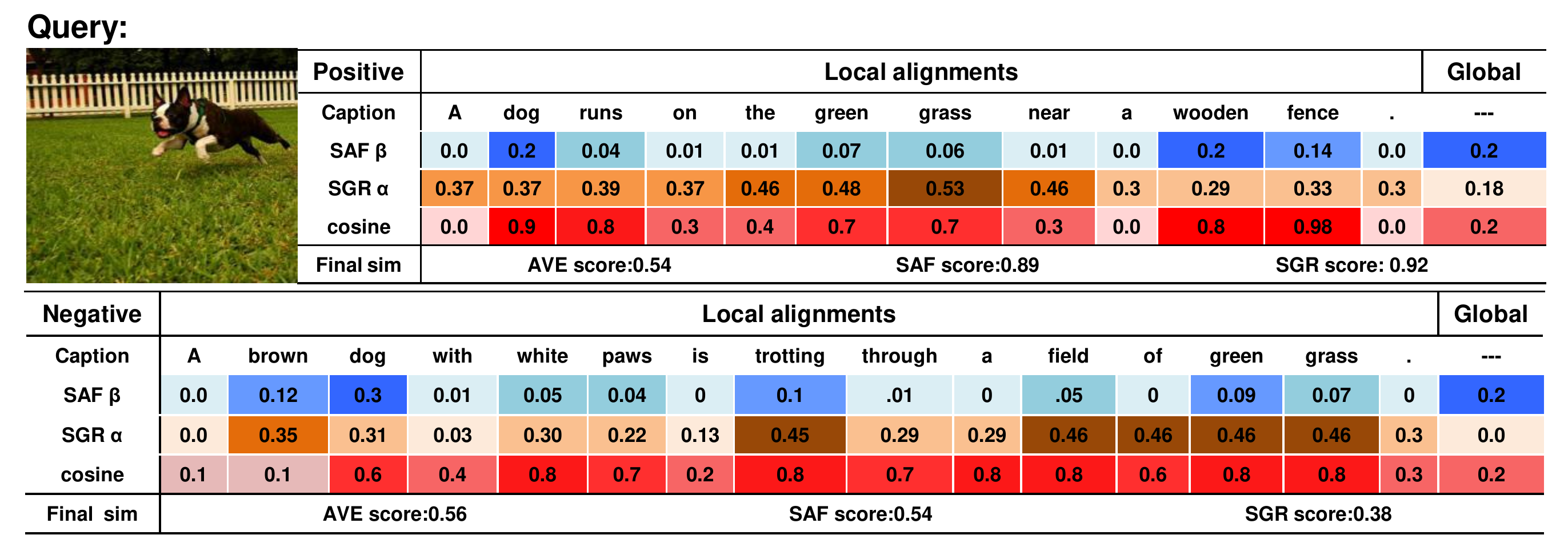} 
	\end{tabular}
	\caption{The visualization of SAF and SGR module. Positive and Negative denotes ground-truth and hard negative examples respectively. SAF $\beta$ denotes attention weight distribution of SAF module. SGR $\alpha$ denotes the cosine distance between final alignment and raw alignments. Final sim denotes similarity calculated by AVE (average), SAF or SGR module} 
	\label{fig:SGRrank}
\end{figure*}

\subsubsection{Configurations for Similarity Computation.}
Table \ref{tabFV} illustrates the impact of different strategies in similarity representation computation and the similarity score prediction. We test the results on local alignments and set the reasoning step of the SGR module to 3. we following\cite{SCAN} to explore two types of the cross-attention modes, i.e. I2T  and T2I. Comparing \#1, \#2, \#5 and \#6, we find that averaging the local alignments calculated by a fully-connected layer and sigmoid function leads to better performance than averaging local cosine distance. Comparing \#3 and \#7, it is more reasonable for the SGR module to count on the local alignments attended by word features (T2I) than the ones by region features (I2T). Besides, the SGR module fails to achieve significant improvement on I2T which indicates that the region features are redundant, relatively independent and irregular in order. Therefore, it is difficult for the SGR module to exploit semantic connections compared with word features. In terms of \#4 and \#8, the SAF module achieves impressive progress both in I2T and T2I modes that demonstrates that the SAF module filters and aggregates plenty of discriminative local alignments steadily to improve the precision of image-text matching.

\subsubsection{Configurations for Training Process.}
In table \ref{tabtrain}, we report the results of different training strategies: joint learning and independent learning. Compared with the SGR/SAF module alone, joint learning can help the SAF module improve the performance of sentence retrieval, and also help the SGR module enhance the ability of image retrieval. In terms of independent learning, the SGRAF network gains an exact and impressive promotion. We assume that the SGR module frequently captures several crucial cues by propagating information between local and global alignments and throws out some relatively unimportant interactions. Moreover, the SAF module attempts to gather all the meaningful alignments and eliminates completely irrelevant interactions. Therefore, the global and local alignments for the SAF and SGR modules are seemingly not incompatible resulting in the unobvious improvement. It is worth noting that the SAF module tends to be more susceptible to the hard negative samples than the SGR module because of the high correlation. On the other hand, it is more challenging for the SGR module to resolve the transmission and integration of numerous semantic alignments. As a result, they can cooperate with each other and further achieve more accurate similarity prediction through independent training.

\subsection{Qualitative Results and Analysis}
As it is shown in Figure \ref{fig:SGRrank}, we illustrate the distribution of attention weights learned by the SAF module. Given an image query, the SAF module captures the key cues (\texttt{"dog runs"}, \texttt{"green grass"}, \texttt{"wooden fence"}) for positive image-text pairs, and also highlights the meaningful instances (\texttt{"brown dog"}, \texttt{"white paws"}, \texttt{"trotting"}, \texttt{"green grass"}) for negative pairs. Note that there exists a crucial discrepancy (\texttt{"brown"}) which is submerged by AVE operation between negative text and image that depicts a black and white dog. Compared with the wrong matching of AVE, SAF module can stress on all the useful alignments including unmatched instance (\texttt{"brown"}) and suppress irrelevant interactions (\texttt{"of"}, \texttt{"with"}, \texttt{"is"}, and etc). On the other hand, the process of SGR module reinforces the role of the alignment (\texttt{"brown"}), which leads to lower similarity between hard negative and query image. Our implementation of this paper is publicly available on GitHub at: \href{https://github.com/Paranioar/SGRAF}{https://github.com/Paranioar/SGRAF}.

\section{Conclusion}
In this work, we present a SGRAF network consisting of similarity graph reasoning (SGR) and similarity attention filtration (SAF) module. The SGR module performs multi-step reasoning based on global and local similarity nodes and captures their relations through information propagation, while the SAF module attends more to discriminative and meaningful alignments for similarity aggregation. We demonstrate that it is important to exploit the relationship between local and global alignments, and suppress the disturbances of less-meaningful alignments. Extensive experiments on benchmark datasets show that both SGR and SAF modules can effectively discover the associations between image and text and achieve further improvements when cooperating with each other.

\clearpage
\section{Acknowledgments}
The paper is supported in part by the National Key R$\&$D Program of China under Grant No. 2018AAA0102001 and National Natural Science Foundation of China under Grant No. 61725202, U1903215, 61829102, 91538201, 61771088, 61751212 and the Fundamental Research Funds for the Central Universities under Grant No. DUT19GJ201 and Dalian Innovation Leader’s Support Plan under Grant No. 2018RD07.

\bibliography{SGRAF}
\clearpage

\begin{appendix}
	\section*{Appendix Overview}
	This supplementary document for similarity reasoning and filtration is organized as follows: 1) more diagrams and descriptions of the SGRAF network: self-attention and SGR module; 2) more quantitative studies: the impact of graph dimension; 3) more qualitative studies: retrieval examples of bidirectional retrieval and visualization of our model. 

	\subsection{Network Details}
	Table \ref{tabSGRAF} shows the detailed implementations of the proposed SGRAF network including generic representation extraction, similarity representation learning, similarity graph reasoning and attention filtration.
	
\begin{figure}[ht]
		\centering
		\begin{tabular}{@{}c}
			\includegraphics[width=0.38\linewidth, height=0.88\linewidth, trim=0 160 800 0,clip]{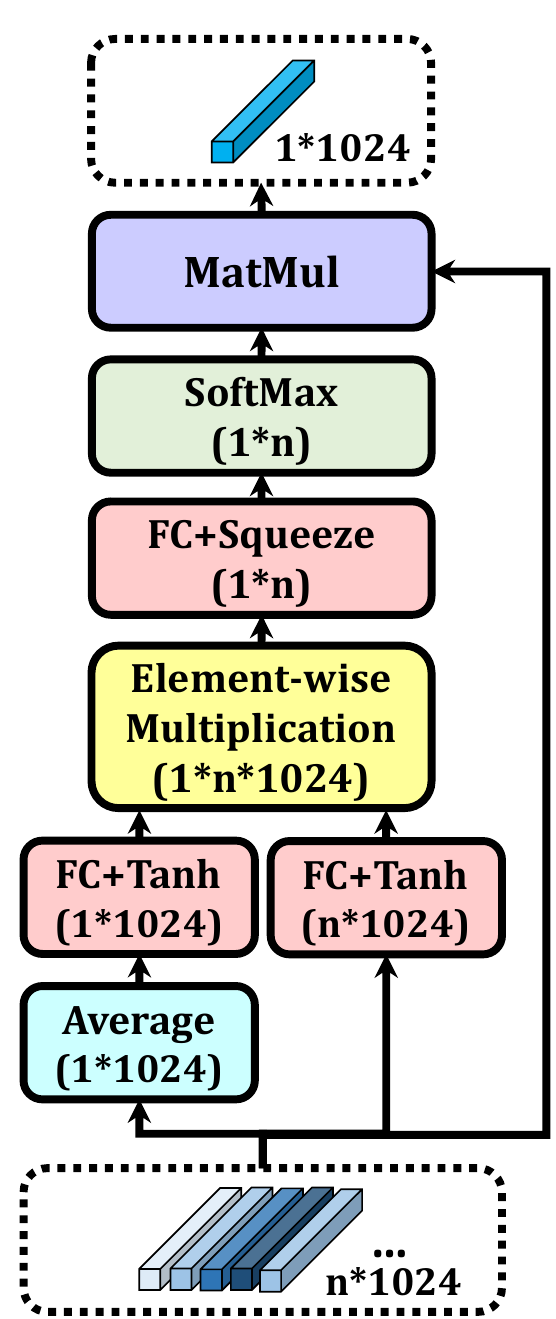}
		\end{tabular}
		\caption{The self-attention module for global representation extraction. $n$ denotes the number of local features, that is, $n = 36$ for image regions and $n = L$ for sentence words}
		\label{fig:SAdetail}
	\end{figure}

	\textbf{Generic Representation Extraction.}
	Given an image, we first apply Faster R-CNN \cite{BU_TDA} to extract the top $K$=36 region proposals and obtain 2048-d feature for each region, then we add a FC layer to transform region features into 1024-d vectors $\boldsymbol{V}$, and perform the self-attention mechanism \cite{Self-Attention} to output a 1024-d global visual vector $\overline{\boldsymbol{v}}$.
	Given a sentence with $L$ words, we transform each word into a 300-d vector with word-embedding, and use Bi-GRU to encode words into 1024-d vectors $\boldsymbol{T}$. Similarly, we exploit the self-attention mechanism \cite{Self-Attention} illustrated in Figure \ref{fig:SAdetail} to output a 1024-d global textual vector $\overline{\boldsymbol{t}}$.

    \textbf{Similarity Representation Learning.}
    We compute $L$ textual-attended 256-d similarity vectors $\boldsymbol{s}^{l}$ with Eq.(5), and one global similarity vector $\boldsymbol{s}^{g}$ with Eq.(2), which obtain $L$+1 (local+global) 256-d similarity vectors $\mathcal{N}$.

\begin{figure*}[ht]
		\centering
		\begin{tabular}{@{}c}
			\includegraphics[width=0.9\linewidth, height=0.37\linewidth, trim=0 250 230 0,clip]{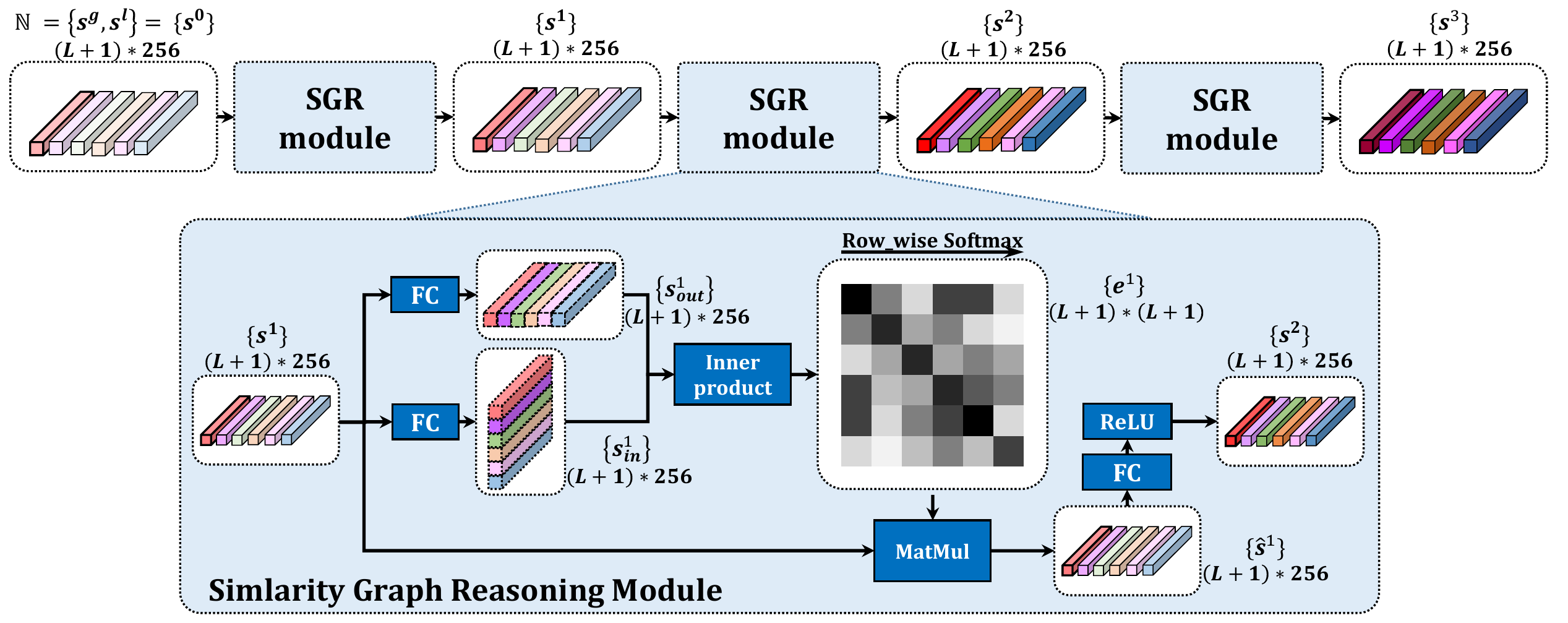}
		\end{tabular}
		\caption{The proposed SGR module for image-text matching. All local alignments $\{\boldsymbol{s}^l\}$ and the global alignment $\{\boldsymbol{s}^g\}$ are firstly taken as initial graph nodes $\{\boldsymbol{s}^0\}$. We compute the edge from node $\boldsymbol{s}_{q} \in \{\boldsymbol{s}^0\}$ to $\boldsymbol{s}_{p} \in \{\boldsymbol{s}^0\}$ by the inner product between incoming and outgoing representations $<\boldsymbol{s}_{in, p}^0, \boldsymbol{s}_{out, q}^0>$, followed by a row-wise softmax. Then the node $\boldsymbol{s}_{p}$ is updated by aggregating its connected nodes (including itself). We iteratively reason the similarity for 3 steps, and take the global node $\boldsymbol{s}^g \in \{\boldsymbol{s}^3\}$ as the reasoned similarity representation}
		\label{fig:SGRdetail}
	\end{figure*}

    \textbf{Similarity Graph Reasoning.}
    As shown in Figure \ref{fig:SGRdetail}, we take the above-introduced $L$+1 (256-d) similarity vectors $\mathcal{N}$ as graph nodes, and then compute the weight of each edge via Eq.(6) with learnable parameter matrices. Graph reasoning is conducted with Eq.(7-8), which means that, for each node $\boldsymbol{s}_{p}$ at step $n$, we learn the weight of its connected nodes (including itself) to aggregate their features from step $n$-1, and then perform a non-linear transformation to update the feature of $\boldsymbol{s}_{p}$ at step $n$. In this way, the information from both local and global alignments is aggregated to produce more accurate similarity predictions. Then we feed the reasoned 256-d global vector $\boldsymbol{s}_{r}$ into a FC+sigmoid layer to output a scalar similarity.
    
    \setlength{\tabcolsep}{2.5pt}
	\begin{table}[ht]
    	\begin{center}
    		\begin{tabular}{c|c|c|c|c}
    			\hline
    			Index &Input &Operation &Symbol &Output\\
    			\hline 
    			\multicolumn{5}{c}{\textbf{Generic Representation Extraction}}\\
    			\hline
    			[1] &(Image) &Faster R-CNN& &36$\times$2048\\
    			\hline
    			[2] &[1] &FC &$\boldsymbol{V}$ &36$\times$1024\\
    			\hline
    			[3] &[2] &Self attention &$\overline{\boldsymbol{v}}$ &1$\times$1024\\
    			\hline
    			[4] &(Sentence) &Word embedding & &$L\times$300\\
    			\hline
    			[5] &[4] &Bi-GRU &$\boldsymbol{T}$ &$L\times$1024\\
    			\hline
    			[6] &[5] &Self attention &$\overline{\boldsymbol{t}}$ &1$\times$1024\\
    			\hline
    			\multicolumn{5}{c}{\textbf{Similarity Representation Learning}}\\
    			\hline
    			[7] &[3],[6] &Eq.(2) &$\boldsymbol{s}^{g}$ &1$\times$256\\
    			\hline
    			[8] &[2],[5] &Eq.(3-5) &$\boldsymbol{s}^{l}$ &$L\times$256\\
    			\hline
    			[9] &[7],[8] &Concatenation &$\mathcal{N}$ &($L$+1)$\times$256\\
    			\hline
    			\multicolumn{5}{c}{\textbf{Similarity Graph Reasoning}}\\
    			\hline
    			[10] &[9] &Eq.(6-8) &&($L$+1)$\times$256\\
    			\hline
    			[11] &[10] &Eq.(6-8) &&($L$+1)$\times$256\\
    			\hline
    			[12] &[11] &Eq.(6-8) &$\boldsymbol{s}_{r}$ &1$\times$256\\
    			\hline
    			[13] &[12] &FC+Sigmoid &&1\\
    			\hline
    			\multicolumn{5}{c}{\textbf{Similarity Attention Filtration}}\\
    			\hline
    			[14] &[9] &Eq.(9) &${\beta}$ &($L$+1)$\times$1\\
    			\hline
    			[15] &[9],[14] &Weighted sum &$\boldsymbol{s}_{f}$ &1$\times$256\\
    			\hline
    			[16] &[15] &FC+Sigmoid &&1\\
    			\hline
    		\end{tabular}
    	\caption{The details of the SGRAF network. $L$ represents the number of words in a sentence, and also denotes the number of local alignments attended by textual words }\label{tabSGRAF}
    	\end{center}
    \end{table}
    
    \textbf{Similarity Attention Filtration.}
    The SAF module takes the $L$+1 (256-d) similarity vectors $\mathcal{N}$ as inputs to learn $L$+1 attention weights $\beta$ with Eq.(9) and performs aggregation to output one 256-d similarity vector $\boldsymbol{s}_{f}$, which is then fed into another FC+sigmoid layer to output a scalar similarity.
    
	\subsection{Quantitative Studies}
	We evaluate the SGR module with different graph dimension $m$ as illustrated in Table \ref{tabdim}. We test the results on global and local alignments and set the reasoning step to 3. The parameters during each step are not shared. We observe that the SGR module is insensitive to the dimension of similarity representation that implies the stabilization and robustness of the SGR module. Note that we set graph dimension $m$ to 256, which can yield the best results for image-text retrieval.
    
    \setlength{\tabcolsep}{3pt}
	\begin{table}[ht]
		\begin{center}
			\renewcommand{\arraystretch}{1.1}
			\begin{tabular}{ccccccc}
				\hline
				\multirow{2}{*}{Graph dim. $m$} & \multicolumn{3}{c}  {Sentence Retrieval} & \multicolumn{3}{c}  {Image Retrieval} 	\\
				\;  & R@1 & R@5   &  R@10  &  R@1 & R@5   &  R@10  \\
				\hline
				128 &74.5 &93.1 &96.3 &54.8 &80.2 &86.2\\
				256 &{\bf75.2} &{\bf93.3} &{\bf96.6} &{\bf56.2} &{\bf81.0} &{\bf86.5}\\
				384 &\bf{75.2} &91.8 &95.5 &54.6 &79.5 &85.6\\
				\hline
			\end{tabular}
			\caption{The impact of Graph Dimension on Flickr30K}\label{tabdim}
		\end{center}
	\end{table}
	
	\subsection{Qualitative Studies}
\begin{figure*}[ht]
		\centering
		\begin{tabular}{@{}c}
			\includegraphics[width=0.90\linewidth, height=0.40\linewidth, trim=0 180 180 0,clip]{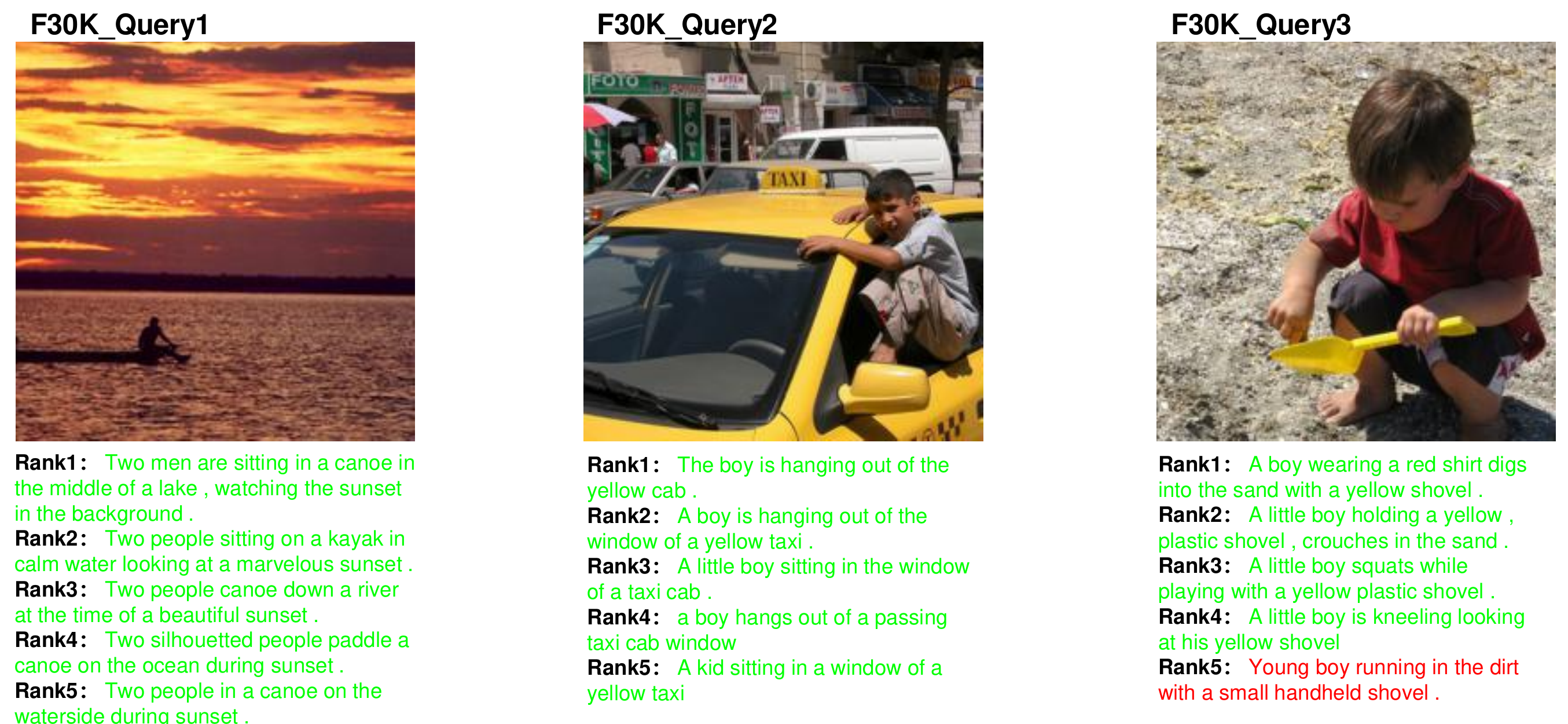}\\\\
			\includegraphics[width=0.90\linewidth, height=0.35\linewidth, trim=0 200 120 0,clip]{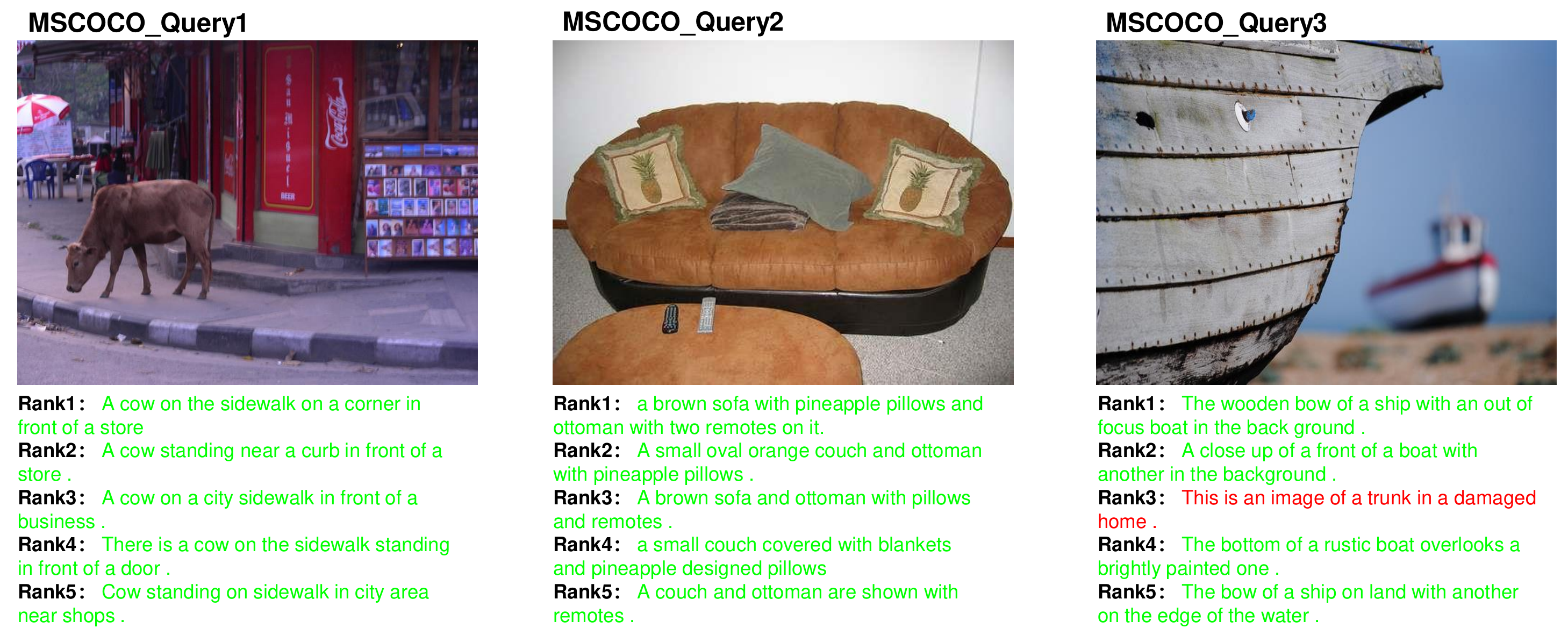}
		\end{tabular}
		\caption{Additional qualitative examples of sentence retrieval on Flickr30K (top) and MSCOCO (bottom). The top-5 retrieved results are displayed. Green denotes the ground-truth sentence and red denotes the unmatched retrieval}
		\label{fig:i2t}
	\end{figure*}

\begin{figure*}[ht]
		\centering
		\begin{tabular}{@{}c}
			\includegraphics[width=0.90\linewidth, height=0.61\linewidth, trim=0 30 220 0,clip]{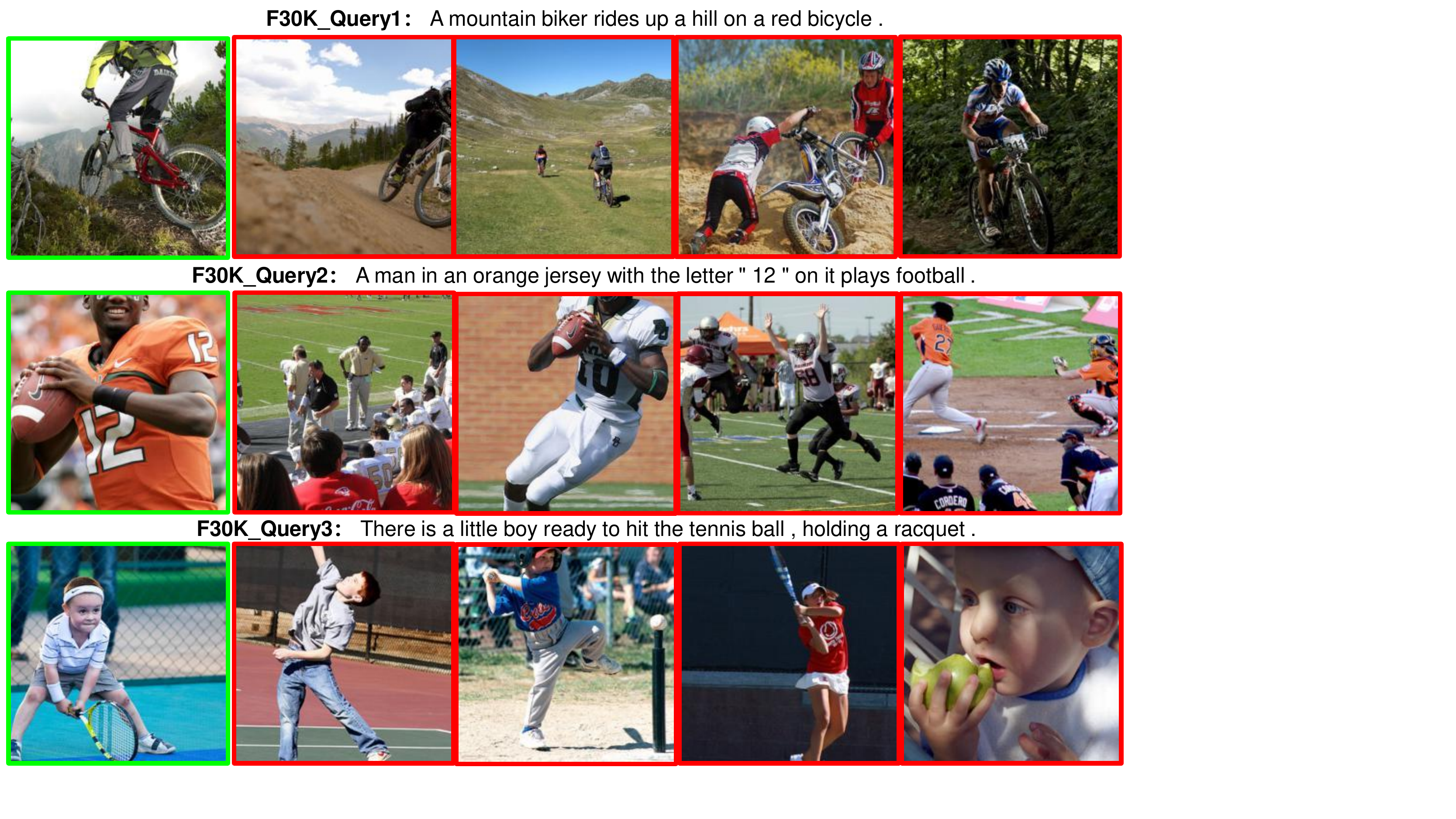}\\\\
			\includegraphics[width=0.90\linewidth, height=0.61\linewidth, trim=0 30 220 0,clip]{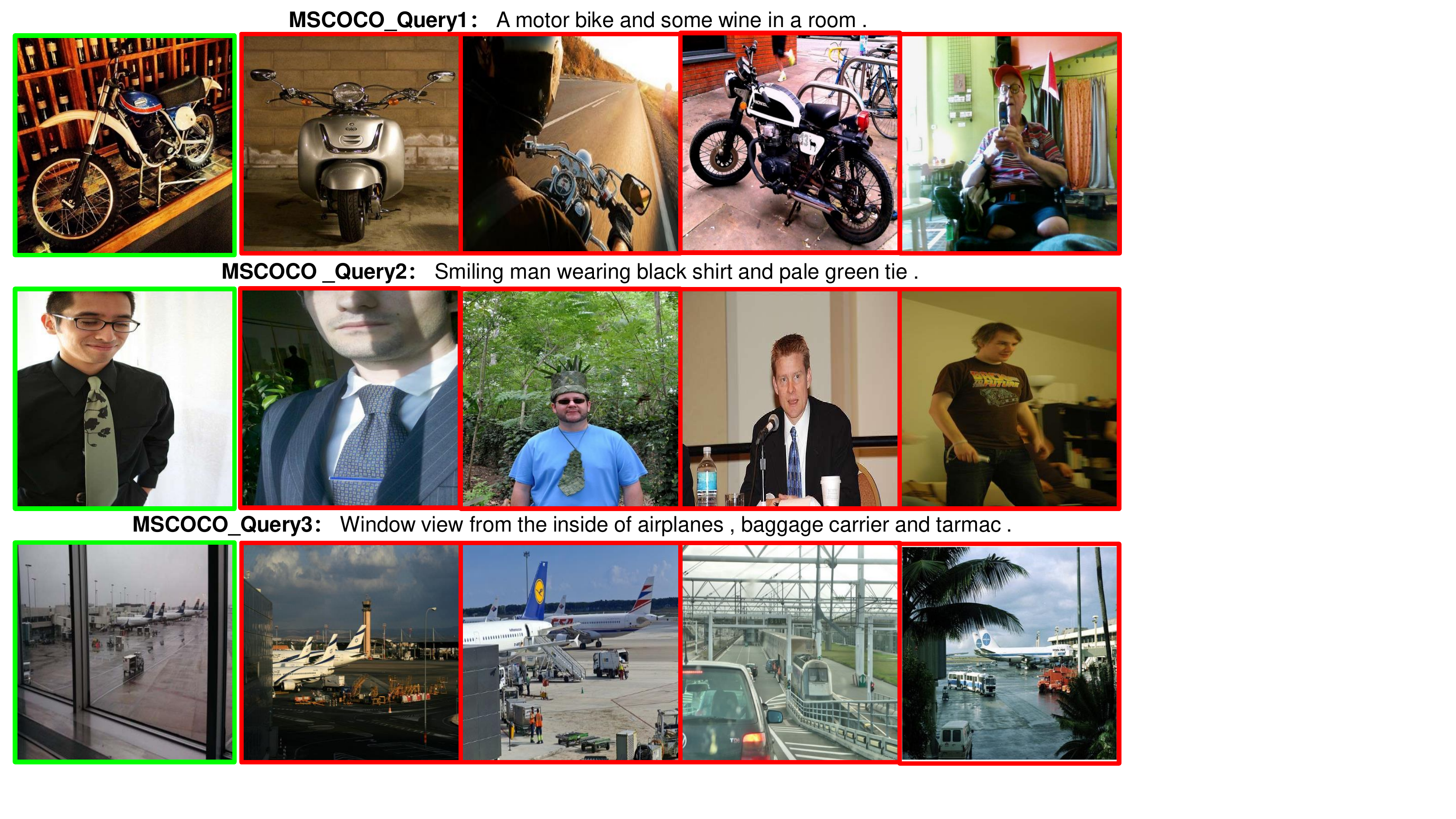}
		\end{tabular}
		\caption{Additional qualitative examples of image retrieval on Flickr30K (top) and MSCOCO (bottom). The top-5 retrieved results are displayed. Green denotes the ground-truth image and red denotes the unmatched retrieval}
		\label{fig:t2i}
	\end{figure*}

\begin{figure*}[ht]
		\centering
		\begin{tabular}{@{}c}
			\includegraphics[width=0.93\linewidth, height=0.27\linewidth, trim=0 290 120 0,clip]{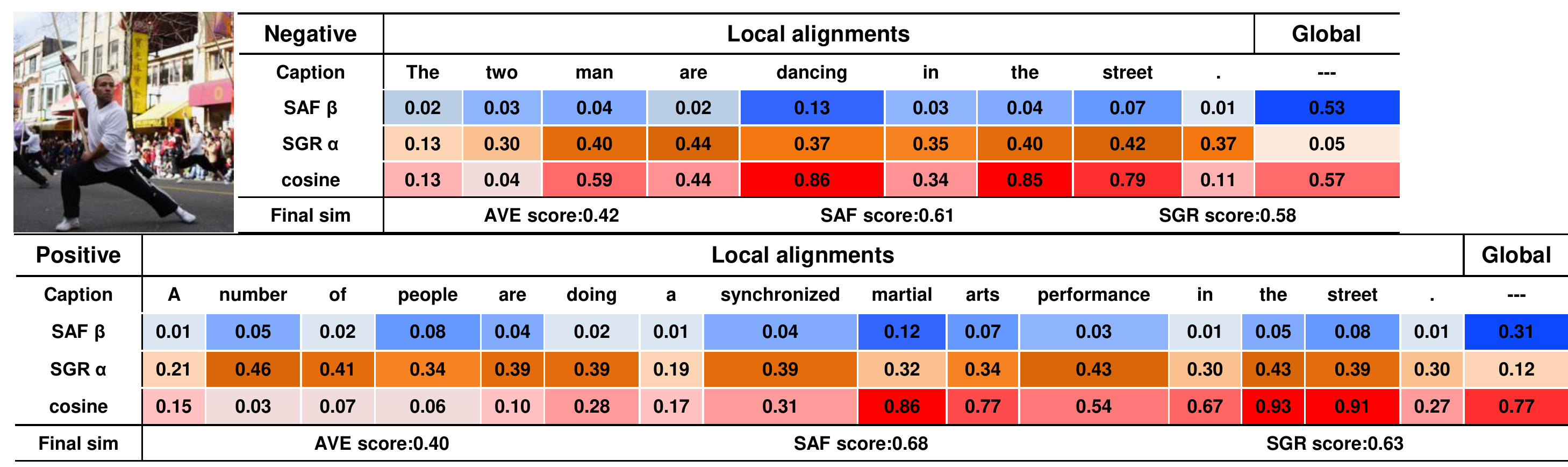}\\\\
			\includegraphics[width=0.98\linewidth, height=0.28\linewidth, trim=0 290 100 0,clip]{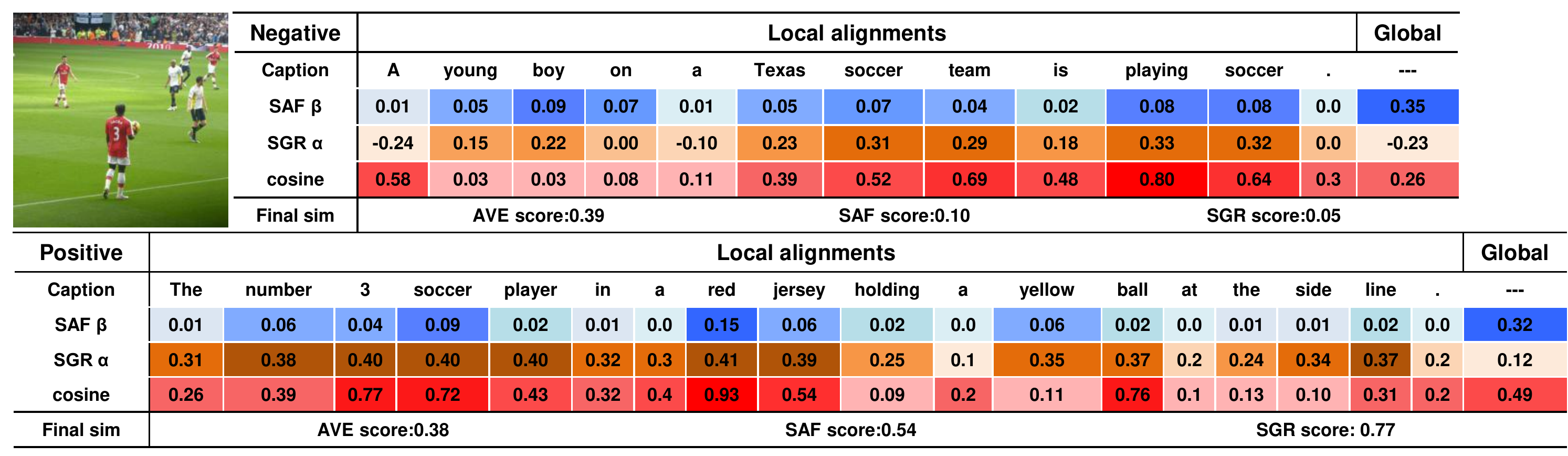} 
		\end{tabular}
		\caption{The visualization of the SGRAF model. Positive and Negative denotes ground-truth and hard negative example respectively. SAF $\beta$ denotes attention weight distribution of SAF module. SGR $\alpha$ denotes the cosine distance between final alignment and raw alignments. Final sim denotes similarity calculated by AVE (average), SAF or SGR module. The key cues of hard negative examples for each query are \{\texttt{"two man"}\} and \{\texttt{"young boy","Texas"}\}. We observe that SAF module can suppress the irrelevant interactions effectively while SGR module can capture fine-grained and crucial alignments by propagating information among all the similarities} 
		\label{fig:weight}
	\end{figure*}

	In this section, we exhibit the retrieval examples of sentence retrieval in Figure \ref{fig:i2t}. Retrieval examples of image retrieval are shown in Figure \ref{fig:t2i}. Furthermore, we demonstrate additional visualization of the SGRAF model in Figure \ref{fig:weight} where the local alignments are attended by textual words.
	
	\textbf{Retrieval Examples of Bidirectional Retrieval.}
	For sentence retrieval, our proposed SGRAF model can efficiently retrieve the correct sentences. Note that the mismatch of F30K-Query3 is also reasonable, which includes highly relevant descriptions of concepts (\texttt{"young boy"}, \texttt{"handheld shovel"}) and scene (\texttt{"dirt"}) with the image.
	For image retrieval, our network can distinguish hard samples well and retrieve the ground-truth image accurately, even if negative samples consist of the same semantic concepts, attributes, and relations with the text descriptions.
	
	\textbf{Visualization of the SGRAF Model.}
	In Figure \ref{fig:weight}, the SAF module can selectively aggregate the discriminative alignments and meanwhile reduce the interferences of less-meaningful alignments, e.g. for the first image query, the SAF module can highlight the key alignments (\texttt{"two man"}, \texttt{"dancing"}, \texttt{"street"}, \texttt{"synchronized martial arts performance"}, etc.) and suppress irrelevant ones (\texttt{"the"}, \texttt{"of"}, \texttt{"in"}, \texttt{"a"}, \texttt{"be"}, etc). Besides, the SGR module can capture fine-grained alignments to achieve comprehensive similarity reasoning, e.g. for the second image query, the SGR module stresses on the alignments (\texttt{"young boy"}, \texttt{"Texas"}) and produces larger gaps between matched and unmatched pairs.
	
\end{appendix}

\end{document}